\documentclass[a4paper,twoside]{article}

\usepackage{amsmath}
\usepackage{amsfonts}
\usepackage{comment}
\usepackage{enumerate}
\usepackage[mathscr]{euscript}
\usepackage{epsfig}
\usepackage{epstopdf}
\usepackage{float}
\usepackage[margin=1in]{geometry}
\usepackage{graphics}
\usepackage{graphicx}
\usepackage{hyperref}
\usepackage[utf8]{inputenc}
\usepackage{latexsym}
\usepackage{mathdots}
\usepackage{pict2e}
\usepackage{subfigure}
\usepackage{setspace}
\usepackage{tikz}
\usepackage{times}
\usepackage{microtype}
\usepackage{subfigure}
\usepackage{epsfig}
\usepackage{subfigure}
\usepackage{calc}
\usepackage{amssymb}
\usepackage{amstext}
\usepackage{amsmath}
\usepackage{amsthm}
\usepackage{wrapfig}
\usepackage{multicol}
\usepackage{pslatex}
\usepackage{apalike}
\usepackage{eufrak}
\usepackage{epstopdf}
\usepackage{caption}

\theoremstyle{definition}
\newtheorem{theorem}{Theorem}[section]

\theoremstyle{definition}

\theoremstyle{remark}
\newtheorem{remark}[theorem]{Remark}

\usepackage{SCITEPRESS}     % Please add other packages that you may need BEFORE the SCITEPRESS.sty package.

\graphicspath{{.}{figs/}}
\DeclareGraphicsExtensions{.pdf,.png,.jpg,.bmp,.eps}

\begin{document}

\title{The Shape of an Image  \subtitle{A Study of Mapper on Images} }

\author{\authorname{Alejandro Robles\sup{1}, Mustafa Hajij\sup{2} and Paul Rosen\sup{2}}
\affiliation{\sup{1}, Department of Electrical Engineering University of South Florida, Tampa, USA}
\affiliation{\sup{2}Department of Computer Science and Engineering, University of South Florida, Tampa, USA}
\email{arobles1@mail.usf.edu,\{mhajij, prosen\}@usf.edu}
}


\abstract{We study the topological construction called \textit{Mapper} in the context of simply connected domains, in particular on images. The Mapper construction can be considered as a generalization for contour, split, and joint trees on simply connected domains. A contour tree on an image domain assumes the height function to be a piecewise linear Morse function. This is a rather restrictive class of functions and does not allow us to explore the topology for most real world images. The Mapper construction avoids this limitation by assuming only continuity on the height function allowing this construction to robustly deal with a significant larger set of images. We provide a customized construction for Mapper on images, give a fast algorithm to compute it, and show how to simplify the Mapper structure in this case. Finally, we provide a simple procedure that guarantees the equivalence of Mapper to contour, join, and split trees on a simply connected domain.}

\onecolumn \maketitle \normalsize \vfill

\section{\uppercase{Introduction}}
\label{sec:introduction}

Recently, the study of data has benefited from the introduction of topological concepts \cite{carlsson2006algebraic,carlsson2009topology,carlsson2008local,carlsson2008persistent,carlsson2005persistence,carlsson2009theory,collins2004barcode}, in a process known as Topological Data Analysis (TDA). 
%Using topology to study data has became to be known now as topological data analysis (TDA). 

One of the most successful topological tools for shape analysis is the \textit{contour tree} \cite{boyell1963hybrid}. The contour tree of a scalar function, defined on a simply connected domain, can be thought of as an efficient topological summary of that domain. This structure is obtained by encoding the evolution of the connectivity of the level sets induced by a scalar function defined on the domain. Reeb  trees are of fundamental importance in computational topology, geometric processing, image processing and computer graphics. 

Contour trees are particularly useful for processing massive data. Contour trees, and their more general version \textit{Reeb graphs} \cite{Reeb1946}, have been used in shape understanding~\cite{attene2003shape}, visualization of isosurfaces \cite{bajaj1997contour}, contour indexing \cite{boyell1963hybrid}, contour extraction \cite{cubes1987high,wyvill1986data}, terrain description \cite{freeman1967searching}, embedding analysis \cite{takeshima2005introducing,zhang2007extraction}, feature detection \cite{takahashi2004topological}, image processing \cite{kweon1994extracting}, data simplification \cite{carr2004simplifying,rosen2017using}, and many other applications. Contour tree algorithms can be found in many papers such as \cite{takahashi2009applying,carr2010flexible,rosen2017hybrid} and Reeb graphs algorithms are studied in \cite{shinagawa1991constructing,cole2003loops,pascucci2007robust,doraiswamy2009efficient}. 

Singh et al.\ proposed a method to understand the shape of data using a topology-inspired construction called \textit{Mapper} \cite{singh2007topological}. Since then, Mapper has became one of the most popular tools used in TDA. It has been applied successfully for various data related problems \cite{lum2013extracting,nicolau2011topology} and studied from multiple points of view \cite{carriere2015structure,dey2017topological,munch2015convergence}.

The construction of Mapper is closely related to Reeb graphs and contour trees \cite{singh2007topological}. Indeed this construction can be considered as a generalization of Reeb graph under some technical conditions \cite{beiwang}. The relation between Reeb graph and Mapper has recently been made precise in \cite{carriere2015structure}.
 
The true power of Mapper lies in its general description in terms of topological spaces and maps on them. This abstract version of the construction is usually called \textit{topological Mapper}. In the original work where Mapper was introduced \cite{singh2007topological}, Mapper was applied to study the shape of point clouds. This version of Mapper is now referred to as \textit{statistical Mapper} \cite{stovner2012mapper}. While topological Mapper allows one to introduce the main ideas of Mapper in general terms, statistical Mapper deals with aspects related to point clouds, such as clustering and noise. 
%that are not addressed directly in the topological version 
Similar technical aspects arise when trying to apply Mappers on other domains, such as images.

The purpose of this article is to study Mapper on specific domains, namely simply connected domains and apply this study to images. While the focus of this article is Mapper on images, we state the results whenever possible on a general simply connected domain.

\subsection{Contribution}

Mapper construction on images operates on a height function defined on the image domain. The height function can be a color channel or luminance of the input image itself or the gradient magnitude of the image, which is typically a compact and connected region in $\mathbb{R}^2$. After discussing the topological and statistical versions of Mapper construction on image domains, we relate this construction to the contour tree algorithm that enables Mapper to realize contour, merge, and split trees.

The method we propose here has multiple advantages. Beside being theoretically justified, the construction of Mapper is flexible and applicable to continuous scalar function defined on a simply connected domain in any dimension. Contour tree algorithms on simply connected domains assume the height function on the domain to be piecewise linear Morse function. While this class of function is useful for a wide variety of applications, it is rather restrictive for images and it does not allow us to explore the topology for most real world images without heavy preprocessing of the image height function. Mapper construction avoids this limitation by assuming only continuity on the height function allowing this construction to robustly deal with a significantly larger class of images. Moreover, Mapper naturally gives a multi-resolution hierarchical understanding of topology of the underlying domain.

The approach we take to Mapper here is geared for simply connected domains and, in particular, for images. Using the properties of such domains, we provide a fast construction algorithm. Finally, we provide a simple algorithm that guarantees the equivalence of Mapper construction to contour, join, and split trees on a simply connected domain.

%The remainder of this paper is organized as follows:

\section{\uppercase{Preliminaries and Motivation}}

As mentioned in the introduction, Mapper is closely related to the contour tree. This related structure motivates the construction of Mapper. 

\begin{figure}[!h]
  %\vspace{-0.2cm}
  \centering
  \hspace{15pt}
   \includegraphics[width=0.97\linewidth]{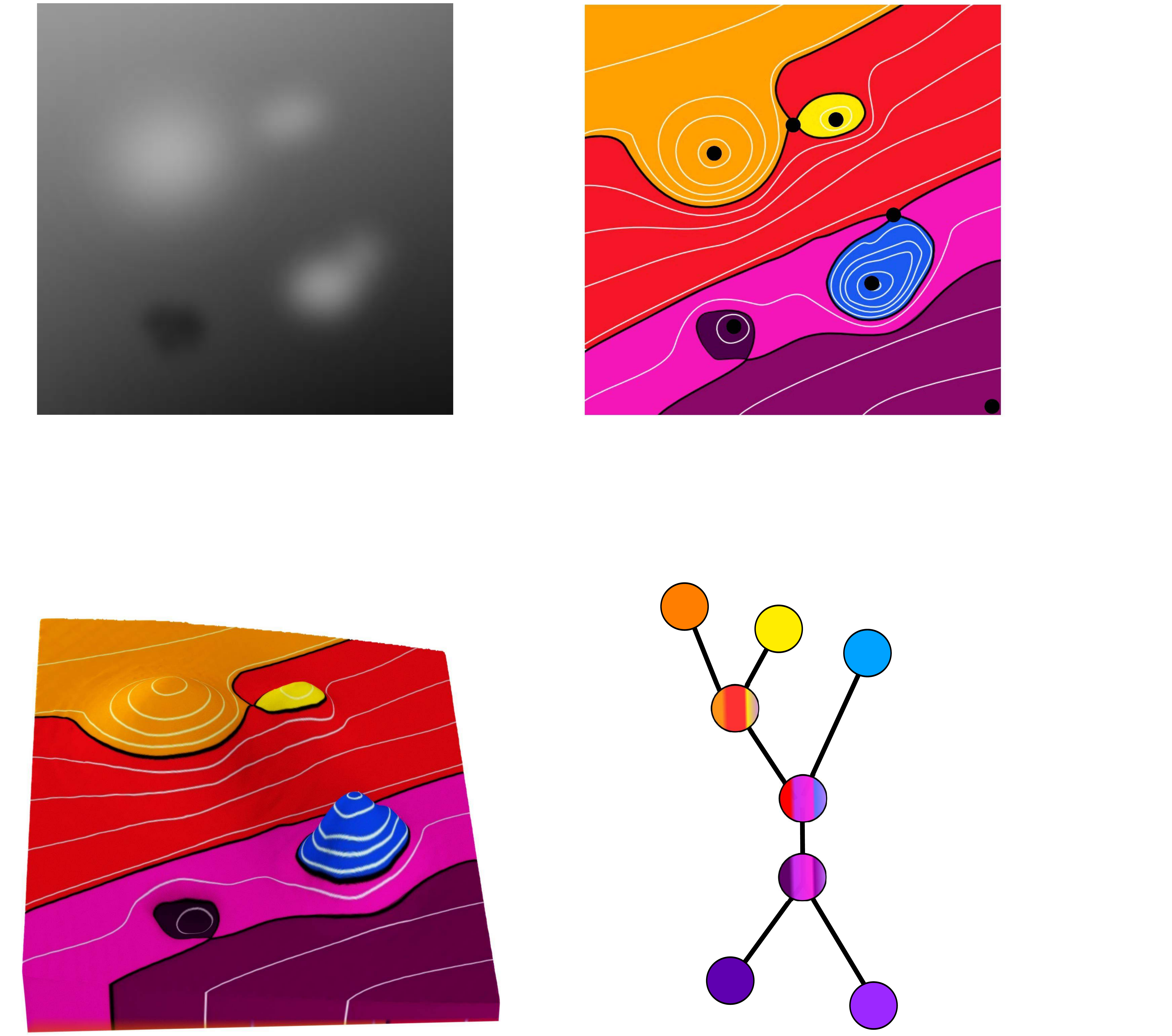}
         \put(-165,-18){(c) }
      \put(-70,85){(b) }
      \put(-165,85){(a) }
      \put(-70,-18){(d) }
  \caption{(a) Scalar function is segmented into (b) topological regions by converting that scalar field into a (c) landscape, using the intensity value for height. The connection of those regions can be converted into a contour tree (d) that describes the topology.}
  \label{fig:c}
 \end{figure}
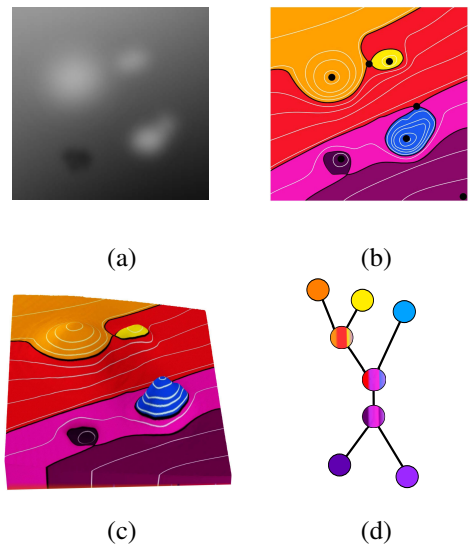

\paragraph{Contour Trees.} The contour tree of a scalar field, defined on a simply connected domain, tracks the evolution of contours in that field and stores this information in a tree structure. Each node in the tree represents a critical point where contours appear, disappear, merge, or split. Each edge corresponds to adjacent and topologically equivalent contours. In essence, the contour tree forms a topological skeleton that connects critical points (i.e.\ local minima, maxima, and saddle points). Figure \ref{fig:c} shows an example of the contour tree of a scalar field defined on a 2d domain.

%One starts with a scalar function $f$ defined on each pixel of an image $X$, say one of the color channel values, and then cover the range of the $f$ with open sets and pull them back to obtain a cover for the  image itself. The graph that one obtains in the case of images is in fact a tree called the \textbf{contour tree} and hence it is simpler than the graphs one possibly can get on a surface. In fact, when $X$ is connected and simple then the Reeb graph is a tree, independent of the function $f$, that is, $R(f)$ is a tree called contour tree.

 %In the case of image $X$ is a compact connected region in $\mathbb{R}^2$.

%In the special case when $X$ is a 2D domain or a 3D domain we obtain what is usually refereed to as 2D and 3D scalar fields.

\begin{figure*}[!t]
\centering
\includegraphics[width=0.95\linewidth]{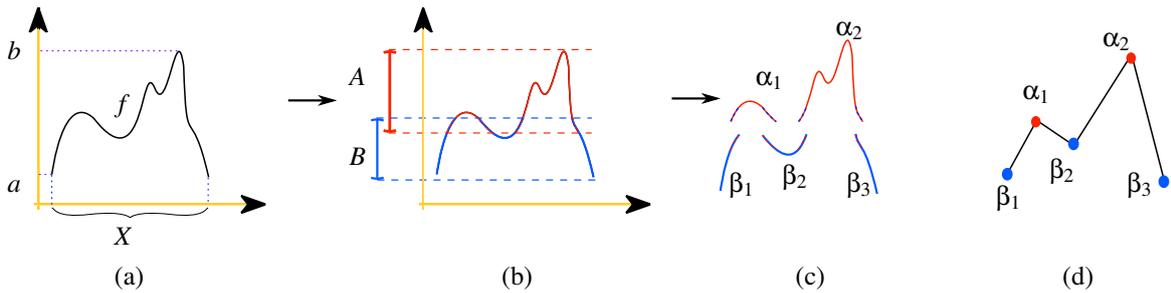}
\put(-435,60){$b$ }
\put(-435,10){$a$ }
\put(-395,-10){$X$}
\put(-395,40){$f$}
\put(-307,20){$B$ }
\put(-307,50){$A$ }
\put(-155,50){$\alpha_1$ }
\put(-125,69){$\alpha_2$ }
\put(-165,10){$\beta_1$ }
\put(-145,13){$\beta_2$ }
\put(-121,10){$\beta_3$ }
\put(-55,45){$\alpha_1$ }
\put(-25,65){$\alpha_2$ }
\put(-65,5){$\beta_1$ }
\put(-45,15){$\beta_2$ }
\put(-16,10){$\beta_3$ }
\put(-395,-25){(a) }
\put(-250,-25){(b) }
\put(-140,-25){(c) }
\put(-40,-25){(d) }

\caption{The construction of Mapper on a 1d function. (a) A scalar function $f:X\longrightarrow [a,b]$. (b) The range $[a,b]$ is covered by the two intervals $A,B$. (c) This gives a decomposition of the domain the domain $X$. The inverse image of $A$ consists of two connected components $\alpha_1$ and $\alpha_2$, and the inverse image of $B$ consists of three connected components $\beta_1$, $\beta_3$ and $\beta_3$. (d) The connected components are represented by the nodes in the Mapper construction. Finally, an edge is inserted whenever two connected components overlap.}
\label{The Mapper construction on a 1d function}
\end{figure*}

Formally speaking, let $X$ be a simply connected domain and let $f:X\longrightarrow [a,b] \subset \mathbb{R}$ be a differentiable scalar function defined on the domain $X$. The nodes of the contour tree of $f$ are represented by the critical points of $f$. Recall that a point $x \in X$ is called a \textit{critical point} of $f$ if the differential $df_x$ is zero. Moreover, a value $c$ in $\mathbb{R}$ is called a \textit{critical value} of the function $f$ is $f^{-1}(c)$ contains a critical point of $f$. On the other hand, if a point in $X$ is not critical then it is called a \textit{regular point}. Similarly, if a value $c \in \mathbb{R}$ is not a critical value then we call it a regular value.

The case when $X$ is an $n$-manifold plays an important role for practical applications. In this case, the inverse function theorem implies that for every regular value $c$ in $\mathbb{R}$ the level set $f^{-1}(c)$ is an $(n-1)$-manifold. For instance, when $X$ is a surface and $c$ is a regular value then $f^{-1}(c)$ is a disjoint union of simply closed curves. If $c$ is a regular value of $f$ then $f^{-1}(c)$ is called an \textit{isosurface}. A \textit{contour} is a connected component of an isosurface. A critical point is called \textit{non-degenerate} if the matrix of the second partial derivatives of $f$ is non-singular. If all the critical points of $f$ are non-degenerate and all critical points have distinct values, then $f$ is a \textit{Morse function} \cite{milnor2016morse}. 

%The \textit{index} of a critical point $x$ of $f$, denoted by $index_f(x)$, is defined to be the number of negative eigenvalues of its Hessian matrix.

%\begin{figure}[h]
%  \centering
%   {\includegraphics[scale=0.2]{minmaxsaddle.jpg}
%   \caption{Minimum, Maximum and Saddle.}
%    \label{minmaxsaddle}
% }
%\end{figure} 

\begin{figure}[!h]
	\centering
	\includegraphics[width=0.4\linewidth]{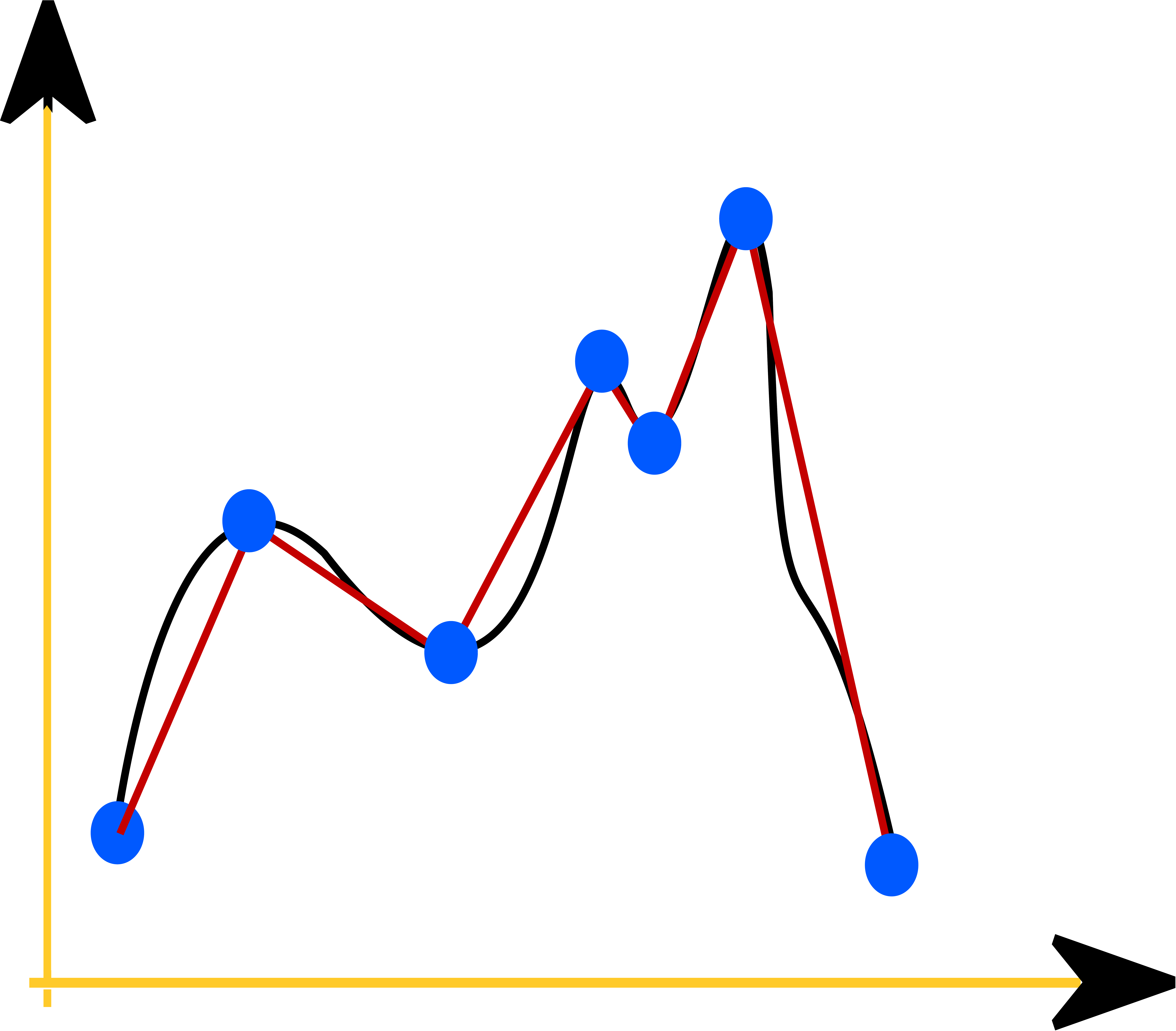}
	\caption{The contour tree of a 1d function.}
	\label{fig:contour tree 1d}
\end{figure}

The contour tree $T(X,f)$ of a Morse scalar function $f$ defined on a simply connected domain $X$ is constructed as follows. Define the equivalence relation $\sim$ on $X$ by $x \sim y$ if and only if $x$ and $y$ belong to the same connected component of a level set $f^{-1}(c)$ for the same $c \in [a,b]$. The set $X/{\sim}$ with the standard quotient topology induced by the function $\pi : X \longrightarrow X/{\sim}$ is called the contour tree of $f$. See Figure \ref{fig:c} for an example of a contour tree defined on the 2d domain of an image. See also Figure \ref{fig:contour tree 1d} for an example the contour tree of 1d function.

In practice, we usually want to compute contour trees on a piecewise linear Morse function defined on a simplicial complex. The mathematical framework specified for contour tree does not apply directly on such domains. The difficulty rises when one tries to extract isosurfaces for a scalar value as the pre-images of an scalar values may not be an isosurface \cite{szymczak2005subdomain}. Nonetheless several contour tree algorithms have been proposed, but they all depend some method of isosurface extraction. Hence two different methods of isosurface extraction might lead to two different contour trees. 

\paragraph{Mapper.}
The construction of Mapper avoids the problem of dealing with isosurfaces all together by focusing on portions of the range of the scalar field. To illustrate this, consider the simple scalar function $f:X\longrightarrow [a,b]$ example given in Figure \ref{The Mapper construction on a 1d function}. Cover the range $[a,b]$ by two overlapping intervals $A:=(a-\epsilon,c+\epsilon)$ and $B=:(c-\epsilon,b+\epsilon)$ such that $c\in [a,b]$ and $\epsilon >0$. Note the interval $A$ and $B$ cover the interval $[a,b]$ in the sense : $[a,b]  \subset  A\cup B$. 

Now, consider the inverse images $f^{-1}(A)$ and $f^{-1}(B)$. Figure \ref{The Mapper construction on a 1d function} (c) illustrates that $f^{-1}(A)$ consists of two connected components $\alpha_1$ and $\alpha_2$ and $f^{-1}(B)$ consists of a three connected components $\beta_1$, $\beta_2$ and $\beta_3$. Moreover, there are some overlaps between these connected components. Namely, the intersections $\alpha_1\cap \beta_1$, $\alpha_1\cap \beta_2$, $\alpha_1\cap \beta_2$ and $\alpha_2\cap \beta_3$ are non-empty. We record the information of the connected components and their non-empty overlap by a graph structure. The nodes of this graph represent the connected components and the edges represent the non-empty intersection between these components. The Mapper construction is the graph associated to the function $f$ and the cover $\{A,B\}$ in this manner. 

\paragraph{Mapper's Relationship to Contour Trees.}
One can notice that this graph is very related to the contour tree of $f$ illustrated in Figure \ref{fig:contour tree 1d}. The only difference in this example seems to be the missing details that the contour tree has but Mapper misses. However, choosing a different cover for the range $[a,b]$ and run the Mapper construction similar to the way we did earlier, one may recover the same structure of the original contour tree.  

\begin{figure}[!h]
  %\vspace{-0.2cm}
  \centering
   {\epsfig{file = 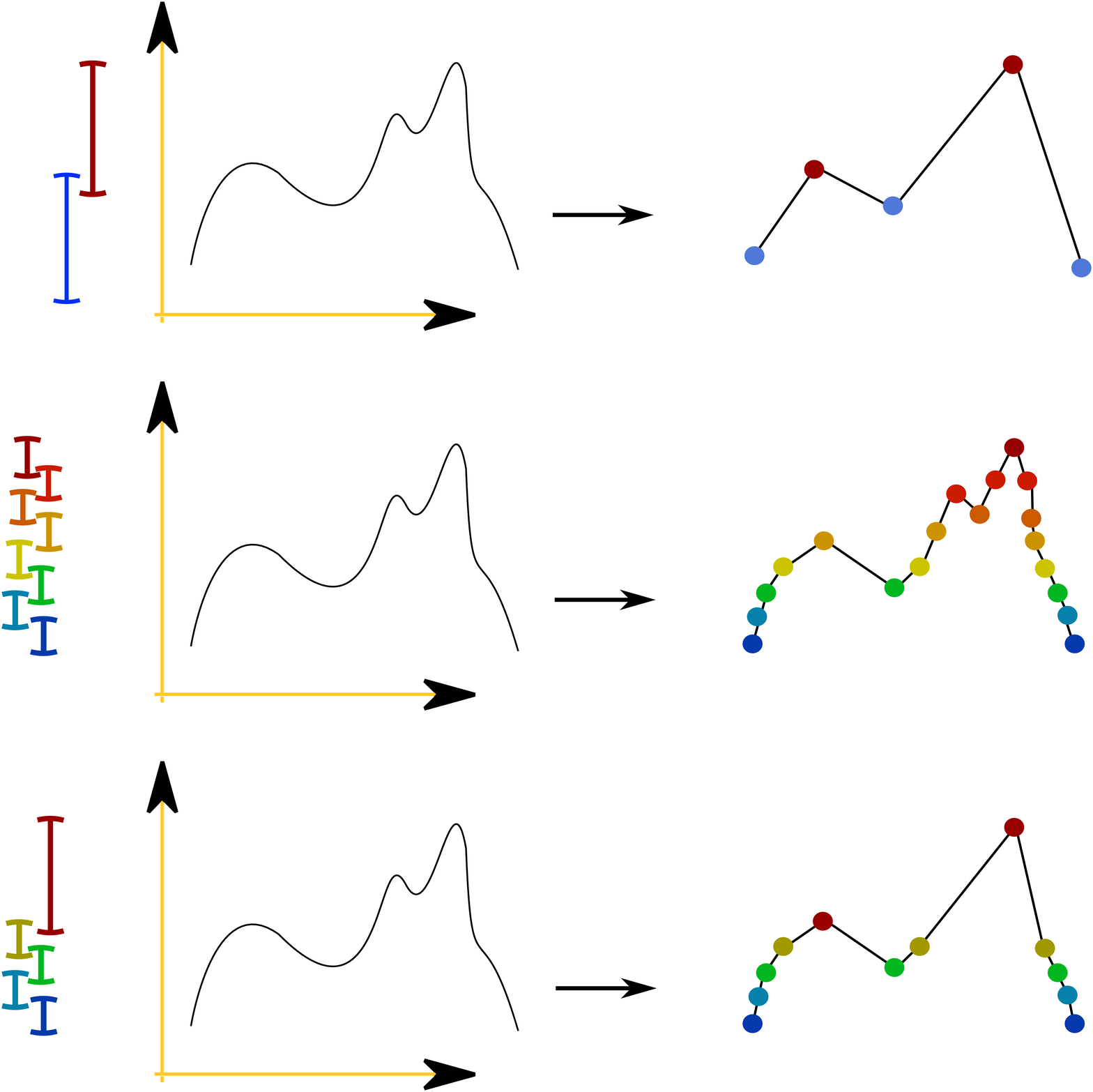, width = 7cm}}
         \put(-90,40){(c) }
      \put(-90,120){(b) }
      \put(-90,190){(a) }
  \caption{The construction of mapper depends on the cover chosen for the range $[a,b]$ of the scalar function. The figure shows three different covers for the range $[a,b]$ and each one gives rise to a different resolution of Mapper.}
  \label{fig:resolutions}
 \end{figure}

The choice of the cover plays an important role in the construction of Mapper, and it allows one to look at the different levels of details of the scalar function and the topology of the considered domain. Figure \ref{fig:resolutions} shows that by increasing the ``resolution'' of the cover imposed on $[a,b]$, one may recover the details encoded in the original contour tree. Note also that the choice of cover given in Figure \ref{fig:resolutions} (b) gives a similar result to the contour tree example given in Figure \ref{fig:contour tree 1d}.

Both contour tree and Mapper essentially track the same topological information in the scalar field, but the way this information is encoded in each one of them is different. The nodes of the contour tree of a scalar field are represented by the critical points the field and the edges represent the regions in the domain where there are no topological change in the contours.  On the other hand the nodes in Mapper represent connected regions in the domain and the edges represent an overlap between two different connected components.

\section{\uppercase{Topological Mapper}}

We now give the general definition of Mapper for a continuous scalar function defined on a simply connected domain.

\begin{figure}[!h]
	\centering
	\includegraphics[width=0.6\linewidth]{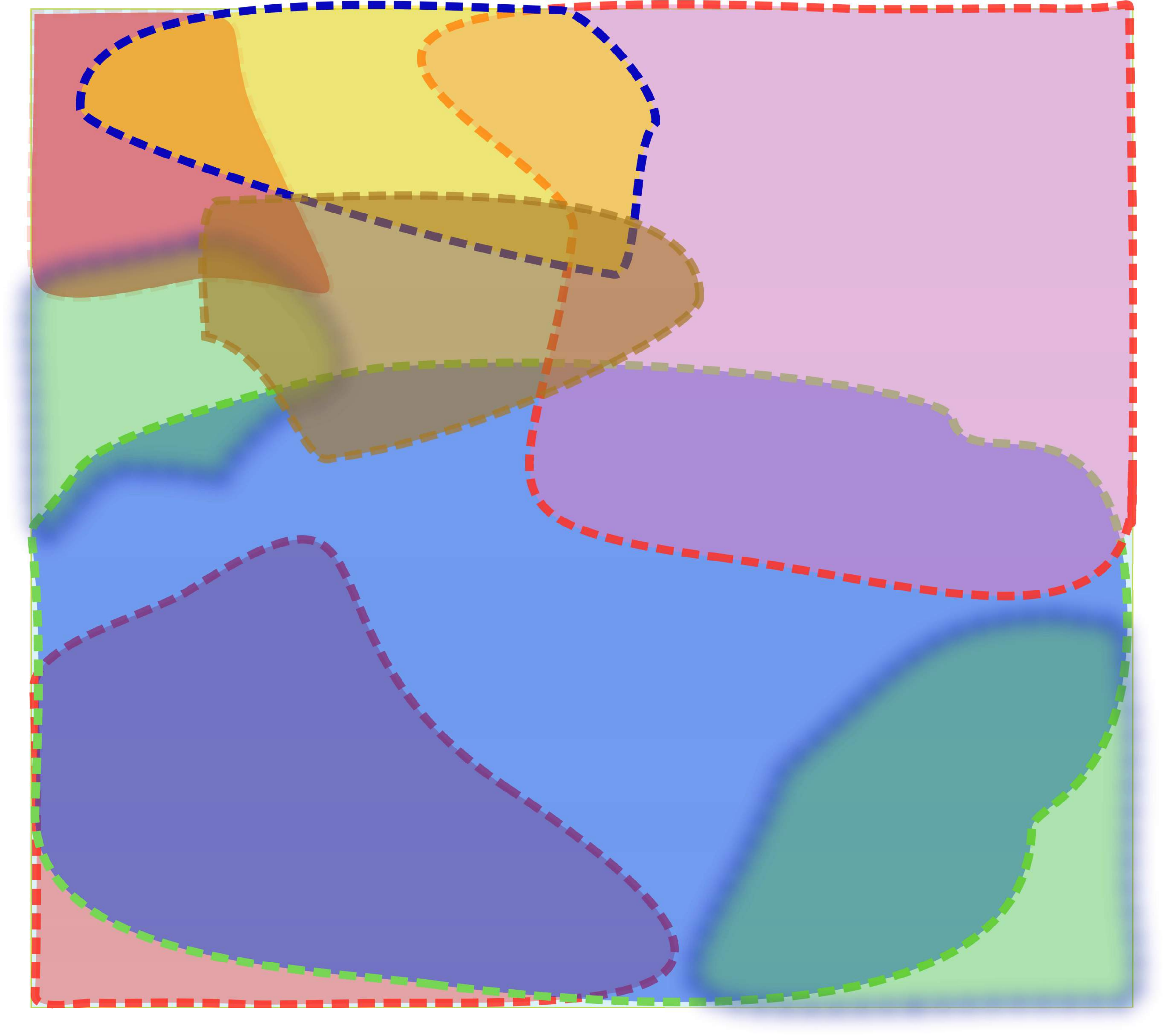}
	\caption{A cover example for a 2d domain.}
	\label{fig:covering example}
\end{figure}

Let $X$ be a simply connected domain in $\mathbb{R}^n$. We will assume that $X$ is compact and connected.  A \textit{cover} of $X$ is a collection of open sets $\mathcal{U}=\{U_i\}_{i\in I}$ such that $X \subset  \cup_{i\in I} U_i$. $I$ here is any indexing set. The compactness condition implies that we can always find a finite cover for $X$. In the case of an image, $X$ is a compact simply connected subset of $\mathbb{R}^2$. See Figure \ref{fig:covering example} for a schematic 2d domain and a cover defined on it.

The $1$-\textbf{nerve} of  $N_1(\mathcal{U,X})$ of $X$ induced by the cover $\mathcal{U}$ is a graph with nodes are represented by the elements of $\mathcal{U}$ and edges represented by the pairs ${A,B}$ of $\mathcal{U}$ such that $A\cap B \neq \emptyset$. The nerve of a space $X$ is well-studied in topology \cite{munkres2000topology}, and it can be thought as a topological skeleton the underlying space. For a general domain $X$, constructing a cover is not a trivial computational task. The main idea of Mapper lies in the way of constructing this cover using the \textit{range} of a function $f$ defined on $X$. The cover of the range can be then \textit{pullback using the function $f$} to obtain a cover for $X$. This cover can be then used to construct the $1$-nerve graph.

More precisely, a continuous scalar function $f:X\longrightarrow [a,b]$ on $X$ and a cover for the range of $f$ give rise to a natural cover of $X$ in the following way. A cover for an interval $[a,b]$ is finite collection of open intervals $\mathcal{U}=\{(a_1,b_1),...,(a_n,b_n)\}$ that cover $[a,b]$, i.e. $[a,b] \subset \cup_{i=1}^n (a_i,b_i)$. Now take the inverse images of each open set in $\mathcal{U}$ under the function $f$. The result is $\mathcal{U}(f):=\{f^{-1}((a_1,b_1)),...,f^{-1}((a_n,b_n))\}$ is an open cover for the space $X$.
The open cover $\mathcal{U}(f)$ can now be used to obtain the $1$-nerve graph $M(X,f,\mathcal{U}):=N_1(X,\mathcal{U}(f))$.  The Mapper construction is by definition the graph $M(X,f,\mathcal{U})$.

\subsection{Cover Resolution}

For a fixed function $f$ the graph $M(X,f,\mathcal{U})$ depends on the choice of the cover $\mathcal{U}$ of the interval $[a,b]$. Figure \ref{fig:resolutions} shows how the choice of the cover affects the Mapper construction. 

This idea of Mapper resolution can be made precise via the notion of  \textit{cover refinement} \cite{munkres2000topology}. Let $X$ be a space and let $\mathcal{A}$ and $\mathcal{B}$ be two covers of $X$. The cover $\mathcal{B}$ is a \textit{refinement} a cover $\mathcal{A}$ if for each element of $B$ of $\mathcal{B}$ there is at least one element $A$ of $\mathcal{A}$ such that $B \subseteq  A$. If $\mathcal{B}$ is a refinement a cover $\mathcal{A}$, there is a embedding of the graph $N_1(X,\mathcal{A})$ inside the graph $N_1(X,\mathcal{B})$. That is there is one-to-one function $\phi$ that maps between the vertices sets $N_1(X,\mathcal{A})$ and $N_1(X,\mathcal{B})$ together with an assignment that assigns to every edge $e=(u,v)$ in $N_1(X,\mathcal{A})$ a path in $N_1(X,\mathcal{B})$ between $\phi(u)$ and $\phi(v)$. See \cite{munkres2000topology}.  

Intuitively, this means that when the resolution of a cover increases, the resulting refined graph obtained from the more refined cover has a copy of the node set of coarse graph. Moreover, each edge of the coarse graph will exist in the refined graph but probably with a higher resolution in the sense that there are some additional nodes inserted along the edge. Figure \ref{result2} show examples $4$ nested sequences of cover refinement along with their corresponding graphs. Starting from left to right, notice how each graph can be embedded in the next graph, in the sense of graph embedding given above. This simple, effective, way to give a multi-resolution Mapper is one of its main advantages over contour tree.

\section{\uppercase{Topological Mapper on Images}}

In this section, we discuss the details of \textit{topological Mapper on images} that will be used in our algorithm for \textit{the statistical Mapper on images} discussed in section \ref{implementation}.

\begin{figure}[!h]
  %\vspace{-0.2cm}
  \centering
   {\epsfig{file = 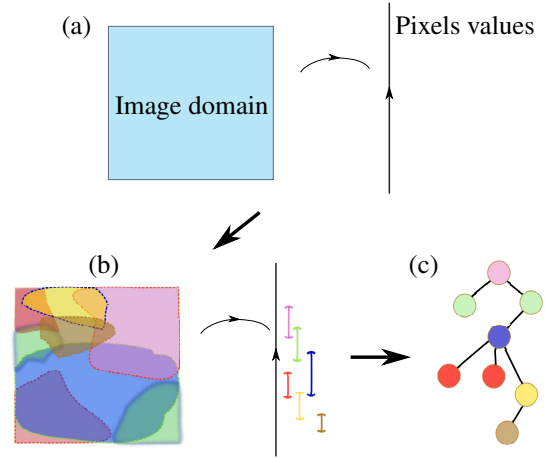, width = 7cm}}
 \put(-180,160){(a) }
 \put(-160,130){Image domain}
  \put(-55,160){Pixels values}
 \put(-170,70){(b) }
 \put(-50,70){(c) }
  \caption{A schematic example of Mapper defined on a 2d domain. (a) A height function is defined on the image domain. (b) Range values of the height function are covered by a collection of open sets and pull them back to the corresponding regions in the image. (c) The Mapper graph is constructed by assigning a node to every connected region in the image and an edge when two regions overlap.}
  \label{fig:mapper 2d}
 \end{figure}

Mapper construction on an image operates on the a height function defined on the domain of the image. The height function can be the gradient magnitude of the image or one of the channels or luminance of the input image. In this section, we assume that $f:X\longrightarrow [a,b]$ is continuous height function defined on the image domain $X \subset \mathbb{R}^2$. The range $[a,b]$ represents the range of possible values for the chosen height function. The idea of Mapper, illustrated previously on $1d$ functions, extends analogously to $2d$ functions. Namely, starting by covering the range $[a,b]$ by a finite collection of open intervals. Then, we find the connected components within the inverse image of each interval and check their intersection. Figure \ref{fig:mapper 2d} shows a schematic example of Mapper on a 2d image domain.

\subsection{Choosing the Cover}
\label{cover}

The choice of cover for the Mapper construction is very flexible. As mentioned in the previous section, this can be used to give a multi-resolution structure that summarizes the scalar function information. That being said, there are certain covers that give rise to a non-desirable tree structure. Moreover, a poor choice of the cover can significantly increase the number of calculations needed for the construction. We describe an effective way to construct the cover for the domain that will help in computing Mapper efficiently. 

Start by splitting the interval $[a,b]$ into n subintervals $[c_1,c_2],[c_2,c_3],...,[c_{n-1},c_{n}]$ such that $c_1=a$ and $c_n=b$. Choose $\epsilon>0$ and construct a cover $\mathcal{U}(\epsilon,n)= \{U_i=(c_{i}-\epsilon,c_{i+1}-\epsilon)\}_{i=1}^{n-1}$ for the interval $[a,b]$. We want to choose $\epsilon$ so that only adjacent intervals intersect. The choice of $\epsilon$ should satisfies the following conditions:

\begin{enumerate}
\item The intersection $U_i \cap U_j =\emptyset $ unless $j\in \{i-1,i,i+1\}$ for $2\leq i,j \leq n-2$.

\item $U_1 \cap U_j =\emptyset $ unless $j\in \{1,2\}$ and finally $U_{n-1}\cap U_j =\emptyset $ unless $j \in \{n-2,n-1\}$.
\end{enumerate}

This choice of $\epsilon$ ensures that only adjacent intervals intersect with each other. Now let $\mathcal{U}_{odd}$ be the subset of $\mathcal{U}$ consisting of intervals with odd indices. Similarly define $\mathcal{U}_{even}$ to be the collection of open sets $U_i \in \mathcal{U} $ such that index $i$ is even. Note that for two open sets $A,B \in \mathcal{U}_{odd}$, we have $A\cap B = \emptyset$. Similarly the intersection of any two sets in $\mathcal{U}_{even}$ is empty. The split of the cover $\mathcal{U}$ in this manner will be utilized in the algorithm. 
%We give more details in section \ref{implementation}.

\subsection{Determining the Nodes}
\label{Nodes}

A node in Mapper is a connected component of $f^{-1}((c,d))$, where $(c,d)$ is an open interval in the cover $\mathcal{U}$ of the range of $f$. Given a range $(c,d)$, in the case of an image $X$, we want to find the those pixels in $X$ whose pixel value lie in $(c,d)$. Given a region $R$ in an image $X$ consisting of a collection of pixels whose pixel value lie within the range $(a,b)$, we want to determine the connected components in the $R$. 

\begin{figure}[!h]
	\centering
	\includegraphics[width=0.95\linewidth]{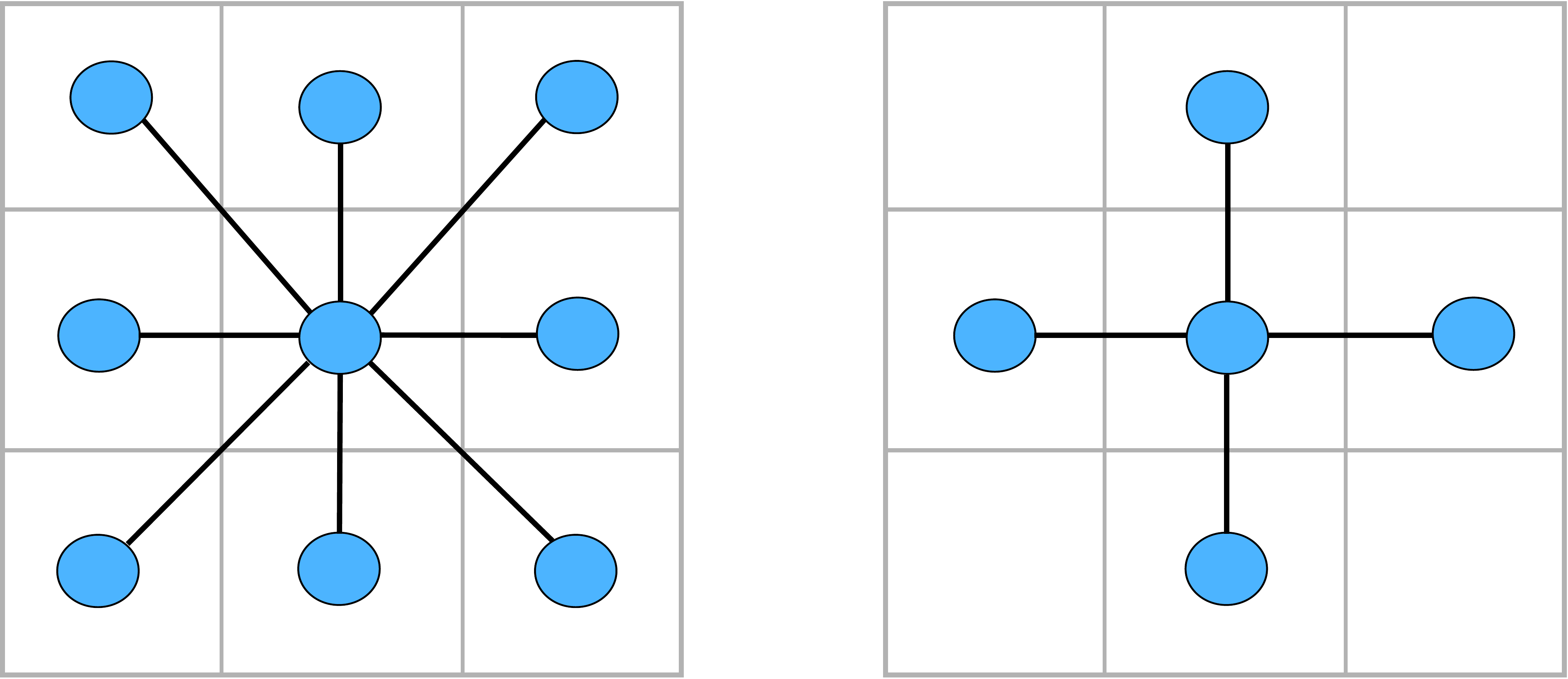}
  \caption{The two types of pixel adjacency relation.}
  \label{fig:local connectivity}
 \end{figure}

Here one needs to specify what exactly is meant by a connected component in this context. The image $X$ induces a graph structure with nodes being the pixels and the edges are determined by the local pixel adjacency relation. There are two common types of pixel adjacency relations shown in Figure \ref{fig:local connectivity}. 

Using the graph on an image with either one of the pixel adjacency relation conventions, we can now consider the connected components of subgraph consists of the pixels in a region $R$. A \emph{walk} on a graph $G$ is a sequence of vertices and edges $(v_0,e_0,v_1,e_1,\cdots,e_{l-1},v_l)$ such that $e_i = [v_{i-1}, v_{i}] \in E(G)$. A graph is said to be \emph{connected} if there is a walk between any two vertices. A connected component in a graph is a maximal connected subgraph. Finding connected components of a graph is well-studied in graph theory and it can be found by in linear time using either breadth-first search or depth-first search \cite{hopcroft1971efficient}. 
%We describe the detail of the way we find the connected components in the case of Mapper in section \ref{implementation}.

\subsection{Determining the Edges}

An edge in Mapper is created whenever two connected components have non-trivial intersection. The cover that we described for the range $[a,b]$ in section \ref{cover} was chosen to minimize the number of sets we check for intersection. Namely the condition that we impose on the cover of $[a,b]$ ensures that only adjacent open interval overlap. In other words, if $U_i$ and $U_j$ are two open sets in the cover of $\mathcal{U}(\epsilon,n)$ of the interval $[a,b]$, then by the choice of the cover specified in section \ref{cover}, we check if the connected components of $f^{-1}(U_i)$ and $f^{-1}(U_j)$ intersect only when we know that $U_i$ and $U_j$ are adjacent to each other. This  significantly reduces the number of set intersections checked. 
%More details are given in section \ref{implementation}.

\section{\uppercase{Algorithm}}
\label{implementation}

The creation of the Mapper graph is done in three stages. First, all pixels in the image are labeled by the cover they map to. Pixels with the same label are then grouped by searching for all connected components with the same label. This provides the nodes for the Mapper graph. Next, the connected component regions are scanned for overlaps. Every pair of overlapping regions in the image corresponds to an edge connecting the nodes in the Mapper graph. Finally, the third stage simplifies the Mapper graph by removing nodes with two valencies.
% This gives us an approximation for Mapper which can then be analyzed.

\subsection{Node Finding}

In our approach, pixel labeling is done using a pair of lookup tables, one for the even cover $\mathcal{U}_{even}$ and one for the odd cover $\mathcal{U}_{odd}$. When a lookup table maps outside of its set of covers, it returns a value that signifies that the pixel does not map to a cover in this  table. This even/odd separation has an important benefit that when one lookup table is applied to the image, none of the resulting regions overlap. This means that instead of processing the image for each cover one-by-one, the image only needs to be processed twice, once for $\mathcal{U}_{even}$ and once for $\mathcal{U}_{odd}$, to find all the connected regions.

Breadth-first search (BFS) is used to find connected regions once the pixels have been labeled. By taking advantage of the queue structure of BFS, every pixel in a connected region can be traversed before moving onto the next region as long as only the top of the queue is being modified. This continuity of the search allows us to add pixels in other regions to the same queue, thus allowing processing many regions with one search. As a region is traversed, pixels are marked with an identification unique to that region. In our implementation, this identification is created using the position of the first pixel in the region touched during the search.

\begin{figure}[!h]
  %\vspace{-0.2cm}
  \centering
   {\epsfig{file = 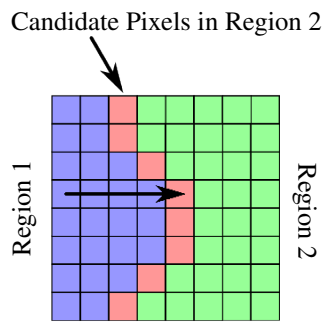, width = 3cm}}
            \put(-100,25){\rotatebox{90}{Region 1} }
            \put(5,60){\rotatebox{-90}{Region 2} }
            \put(-103,110){ Candidate Pixels in Region 2 }
  \caption{Line scanning for candidate pixels}
  \label{fig:line scanning}
 \end{figure}

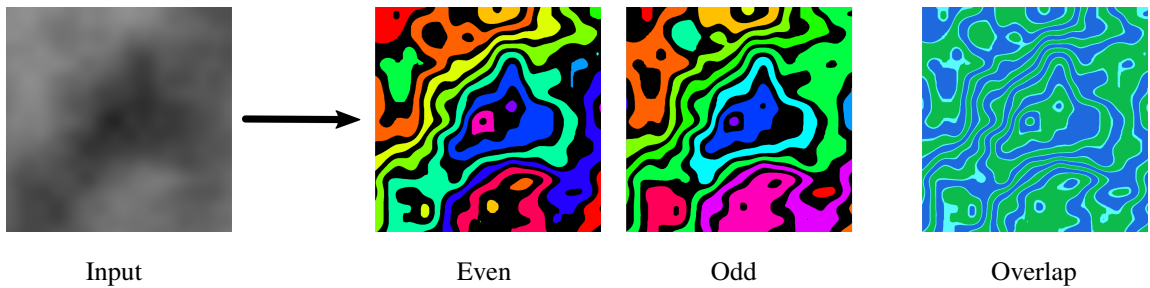
\begin{figure*}[!t]
\centering
\includegraphics[width=0.95\linewidth]{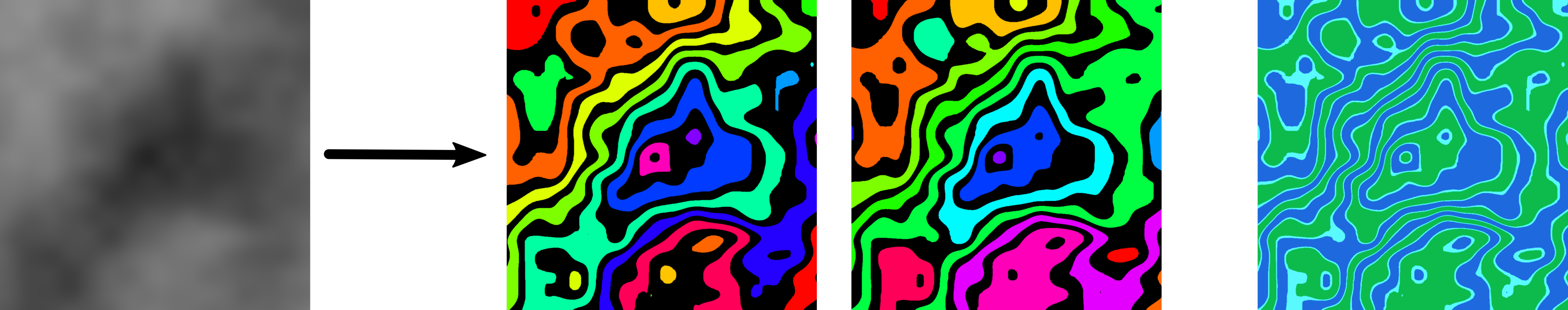}
\put(-400,-18){ Input }
\put(-261,-18){ Even }
\put(-166,-18){ Odd }
\put(-61,-18){ Overlap }
\caption{Region search applied to Perlin noise. The search is done twice, once for even and once for odd covers. Here, each region identified during the search is given a unique color. If a pixel is not found to map to a cover during the search the, pixel is not colored (these are the pixels colored in black in the middle two images). This shows how splitting the covers gives a pair of images which do not contain overlapping regions. Regions in one image will, however, overlap with regions in the other image, as shown in the image on the far left.}
\label{even_odd_split}
\end{figure*}

Our approach initializes the BFS queue with \textit{candidate pixels} which are pixels found by scanning each row in the image from left to right until a pixel which differs in label from the previous pixel is found (see Figure \ref{fig:line scanning}). This gives the pixels which start a region along every line in the image. Since a region needs at least one pixel to be in the queue at the start of the search, the use of candidate pixels ensures each region in the image will be traversed, while reducing the number of pixels in the queue at the start of the search.

At the end of the search, every pixel will have an associated identification that represents the connected component region it belongs to. Finally, these regions define the nodes in the Mapper graph. See Figure \ref{even_odd_split} for illustration of the process of node finding done on an example image.
%Since each region is identified by the location of the first pixel in the region touched during the search, these identifications will be arbitrary. Thus, in order to index these regions, a map has to be created from these identifications to the nodes it represents in the graph. This process needs to be done for both the even and odd coverings.

\subsection{Edge Finding}

Once the regions in the image have been identified for both the even and odd covers, overlaps between regions need to be found. A naive approach would be to create a set of pixel locations for every region in both sets of covers, and check whether pairs of sets are disjoint. This type of approach, however, requires every pair of sets to be tested for disjointness, making it inefficient. 

To determine region overlap, we take advantage of the candidate pixels found during node finding, see Figure \ref{fig:line scanning}. Since these pixels signifies the entrance of a region with a different labeling, this means that there are two different regions from the two opposing covers overlap.  %we use is that only the candidate pixels found during node finding, shown in Figure \ref{fig:line scanning}, are used to find overlapping regions in the opposing set of coverings. 
Notice that this method takes advantage of the way we construct the cover in section \ref{cover}. %and creates a list of edge pairs which contains duplicates of the same pairs. These duplicates are then removed by sorting the list and removing consecutive pairs. An edge in the Mapper graph is determined by such a pair.

\subsection{Graph Simplification}
\label{sec:optimization}

\begin{figure}[!h]
	\centering
	\includegraphics[width=0.975\linewidth]{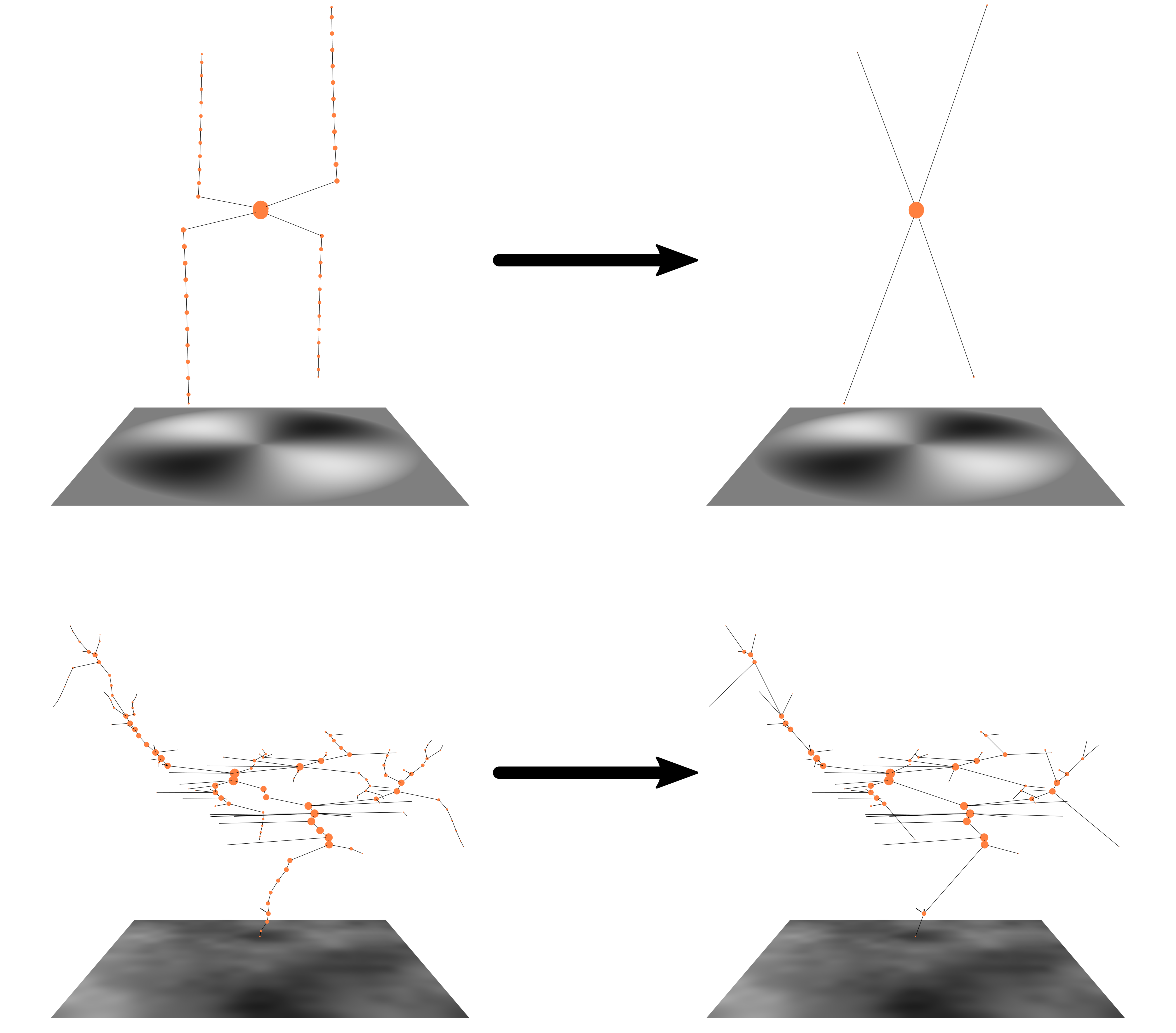}
	\caption{Simplification of the mapper graph for a saddle point and Perlin noise. This shows how information about the topology is retained after the simplification.}
	\label{fig:simplification}
 \end{figure}

The resulting Mapper graph can contain thousands of nodes. Many of these nodes can be removed as they do not indicate topological events. In the Mapper graph, a node with valency equal to $2$ corresponds to a region where no topological event occur. In other words, such a node is not a merge, split, creation, or termination of a region. These nodes are analogous to regular points in the contour tree. Hence, these nodes can be safely removed to obtain a simplified graph, such as in Figure \ref{fig:simplification}.

\section{\uppercase{Realizing the Contour Tree}}

The Mapper construction can be used to realize the contour tree. Here we give a choice of cover that guarantees Mapper gives rise to all the topological information encoded in the contour tree. We need to assume that the given scalar function is a piecewise linear Morse function $f:X\longrightarrow [a,b]$ on a simply connected domain $X$. The assumption of piecewise linear Morse is necessary in order to work with a contour tree. For precise definitions related to Morse theory on simplicial complex see \cite{pascucci2004multi}.

Recall that every node in the contour tree corresponds to a critical point. The critical point of a function signifies a topological change in the space $X$ with respect the scalar function. Moreover, if $t_1$ and $t_2$ are two consecutive critical values of $f$ then for any two values $c_1,c_2 \in (t_1,t_2)$ the number of connected components of both $f^{-1}(c_1)$ and $f^{-1}(c_2)$ are the same. In other words, topological changes occur to a level set $f^{-1}(t)$ only when as $t$ sweeps though a critical value. Hence, in order for Mapper to give us the information encoded in the contour tree, it is sufficient to make a choice of the cover on $[a,b]$, so that we store the following information:

\begin{enumerate}
\item The number of connected components between every two consecutive critical values of $f$.
\item The way the connected components merge, split, appear, and disappear when passing through a critical point.
\end{enumerate}

\noindent
The following procedure gives a choice of cover for $[a,b]$ that satisfies the previous two criteria:
\begin{enumerate}
\item  Let $t_1,t_2,...,t_n$ be the critical values for $f$ ordered in an ascending order. Let $p_1,p_2,...,p_n$ be the corresponding critical points of $f$. 

\item For each $1\leq i\leq n-1$, we choose four numbers $a_i$ $b_i,c_i$ and $c_i$ in the interval $[t_i,t_{i+1}]$ such that $a_i<d_i<c_i<b_i$.

\item Let $c_0=a-\epsilon$ and let $d_n=b+\epsilon$ for some $\epsilon >0$. 

\item Let $\mathcal{U}$ be the cover of $[a,b]$ consisting of the intervals $(a_1,b_1),...,(a_{n-1},b_{n-1})$ as well as $(c_0,d_1)$,$(c_1,d_2)$,...,$(c_{n-1},d_n)$.
\end{enumerate}

Notice that the Mapper construction obtained using the covering $\mathcal{U}$ given above stores all the topological information encoded in the function $f$. Hence, any further refinement of the covering $\mathcal{U}$ will not produce any further details in the Mapper construction as far as the topology of the original domain is concerned. In other words, the above construction gives the highest Mapper resolution that one could obtain on a piecewise linear Morse function. 

\begin{remark}
Notice that the Mapper construction does not need nor assume the function to be a Morse function. However, this assumption is needed in this section because we want to show that in the case when the function is Morse, then Mapper can give essentially the same structure as the contour tree.
\end{remark}

\section{Join and Split Trees}

The previous sections describe how Mapper can be used to obtain a contour tree. The Mapper construction is general and can be used to realize other structures such as the join and split trees \cite{carr2003computing11}. The only change one needs to make to the previous setup is making a different choice for the shape of the open intervals that form the cover of the range. These choices will be justified after we illustrate the basic ideas of join/split trees. 
 
For a continuous scalar function $f:X\longrightarrow [a,b]$ defined on a simply connected domain $X$ the split tree $ST(f,X)$ of $f$ on $X$ tracks the topological changes occur of the set $\{p\in X|f(p)\geq c \}$ of a value $c$ as this value is swept from $\infty$ to $-\infty$. Similarly, the join tree $JT(f,X)$ of $f$ on $X$ tracks the topological changes occur to the topology of the set $\{ p\in X| f(p) \leq c \}$ as the value $c$ goes from $-\infty$ to $\infty$. Note that join and split trees can be obtained from the contour tree. Namely, if one sweeps the contour tree from bottom to top, keeping track of the merging events and ignoring splitting events, then one obtains the join tree. On the other hand, if we sweep the contour tree from top to bottom, keeping track of the splitting events only, then we obtain the split tree. The join and split trees can also be used together to reconstruct the contour tree \cite{carr2003computing11}. The Mapper construction can be used to compute both split and join trees on any simply connected domain. The setup to obtain these two structures is similar to the one we demonstrated for the contour tree. The only difference is the shape of the open intervals for the cover $\mathcal{U}$ of range $[a,b]$.

The choice of cover for a join tree should be of a collection of open intervals of the form $(-\infty,c)$ that covers the interval $[a,b]$. That is, the cover must be a finite set $\{(-\infty,c_1),...,(-\infty,c_n)\}$ such that $[a,b] \subset \cup_{i=1}^n(-\infty,c_i)$. As the values to $c_i$ increase the only merging events occur in the set $\{p\in X|f(p)\geq c \}$, which is reflected in the resulting Mapper graph. 
%In the setup of images, the precious analysis we made in this paper applies with no change.
On the other hand, the choice of cover needed to construct the split tree is a collection of open intervals of the form $(c,\infty)$.

\section{\uppercase{Results}}
\label{sec:results}

\begin{figure*}[!t]
	\centering
	\includegraphics[width=0.95\linewidth]{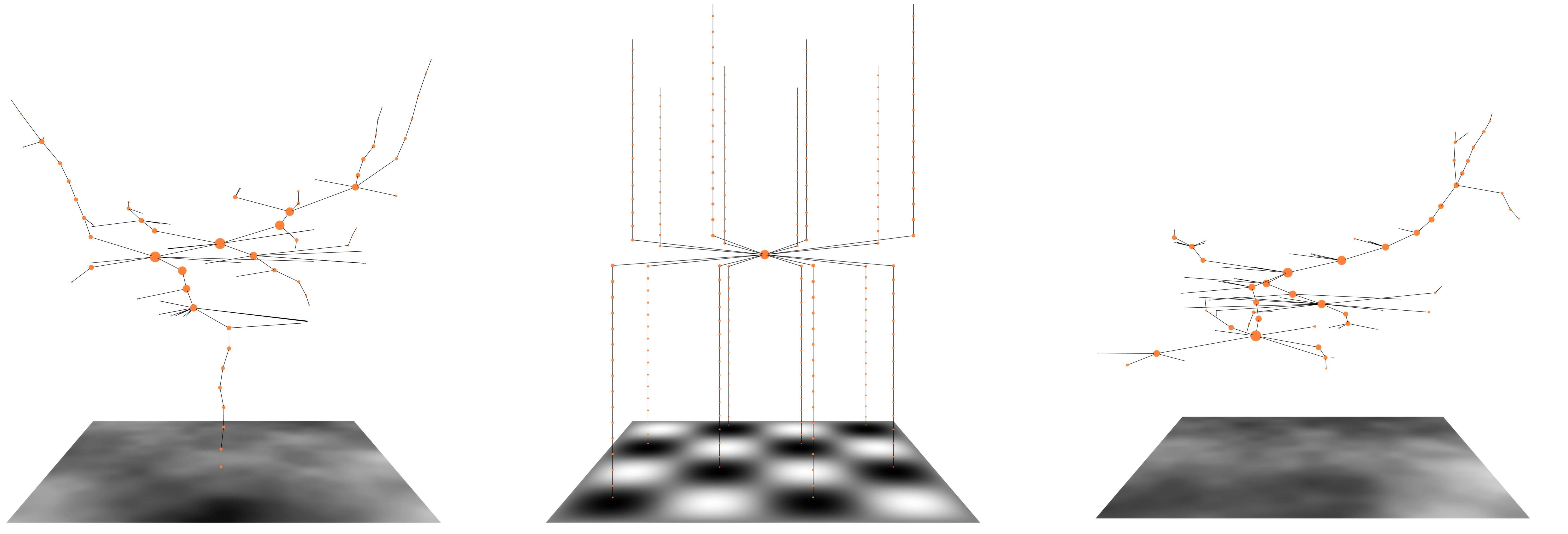}
	\caption{Examples of Mapper on images using pixel values as the height function. The range of these images was covered by a cover of $32$ open sets.}
	\label{result1}
\end{figure*}

To demonstrate how our work performs we run a few experiments on some images with various complexities. Figure \ref{result1} shows the illustrative examples on some images. The height functions chosen on these images are the input images themselves. The figure shows the images along with the Mapper graph on drawn on the top of them. The vertical position of the node is chosen to be the average of the pixel values of the region that corresponds to that node. On the other hand the $(x,y)$ position of a node is the center mass of the pixel positions of the pixels in the region. The size of the node is proportional to the number of pixels in the corresponding connected component.

\begin{figure*}[!t]
	\centering
	\includegraphics[width=0.95\linewidth]{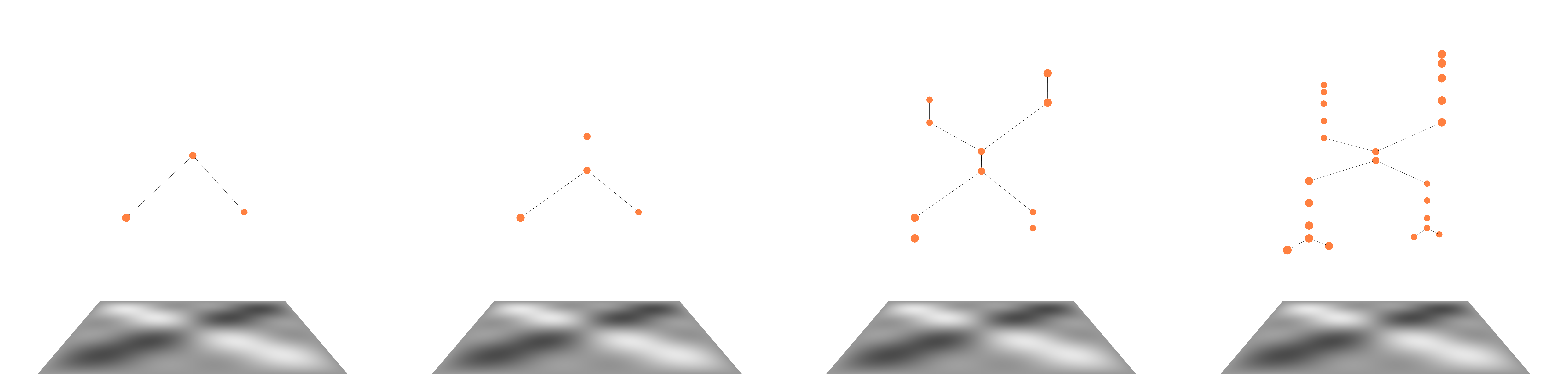}
	\caption{Multi-resolution of Mapper using different cover resolutions. The graphs are constructed from left to right by using $2,4,8,16$ slices of the range cover.}
	\label{result2}
\end{figure*}
\begin{figure*}[!t]
	\centering
	\includegraphics[width=0.95\linewidth]{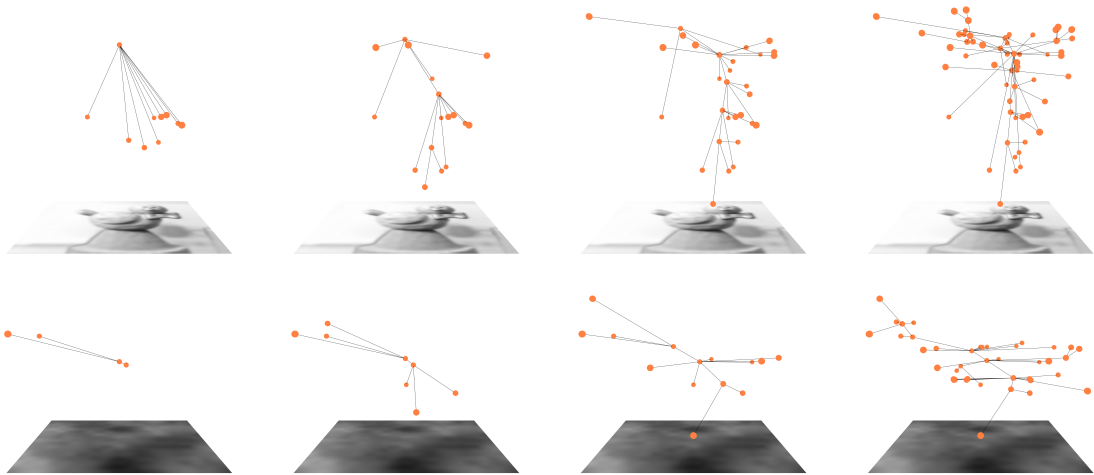}
	\caption{Multi-resolution of Mapper using different cover resolutions. For each image the graphs are constructed from left to right by using $2,4,8,16$ slices of the range cover.}
	\label{result3}
\end{figure*}

In Figure \ref{result2} we show how multiple refinement of cover give rise to a hierarchy of Mapper on the same image. The graphs in the figure, shown from left to right, are generated by using $2,4,8,16$ slices of the cover. The figure shows immediately the effect of cover refinement of the resolution and level of details.

The same hierarchy construction of Mapper is applied to more complicated images and shown in Figure \ref{result3}. For a better visualization, the graphs in this figure were optimized as described in \ref{sec:optimization}.

\begin{figure*}[!t]
	\centering
	\includegraphics[width=1\linewidth]{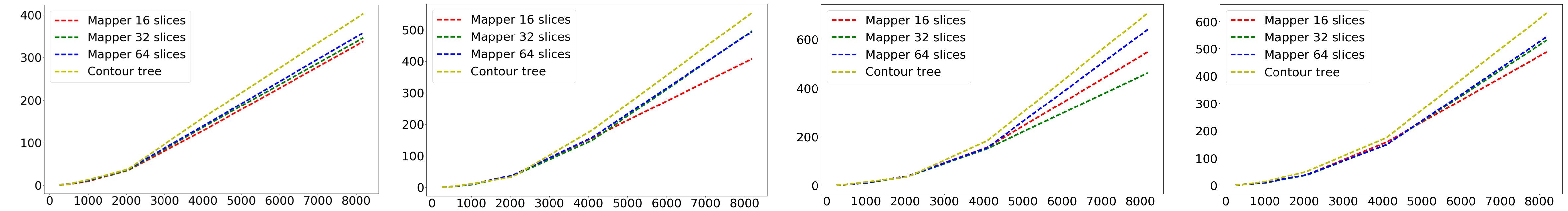}
	\caption{Performance analysis of Mapper in comparison with contour tree on the four patterns given in Figure \ref{fig:patterns}. Tests were done on images with the resolutions : $256^2, 512^2, 1024^2,2084^2,4096^2,$ and $8192^2$. Each resolution was tested against Mapper and contour tree. Mapper was tested using $16,32$ and $64$ cover slices. The $x$-axis represents the square root of the  resolution of the image. The $y$-axis represents the running time in milliseconds. }  
	\label{performance}
\end{figure*}

\subsection{Running time}
\begin{figure}[!h]
	\centering
	\includegraphics[width=0.8\linewidth]{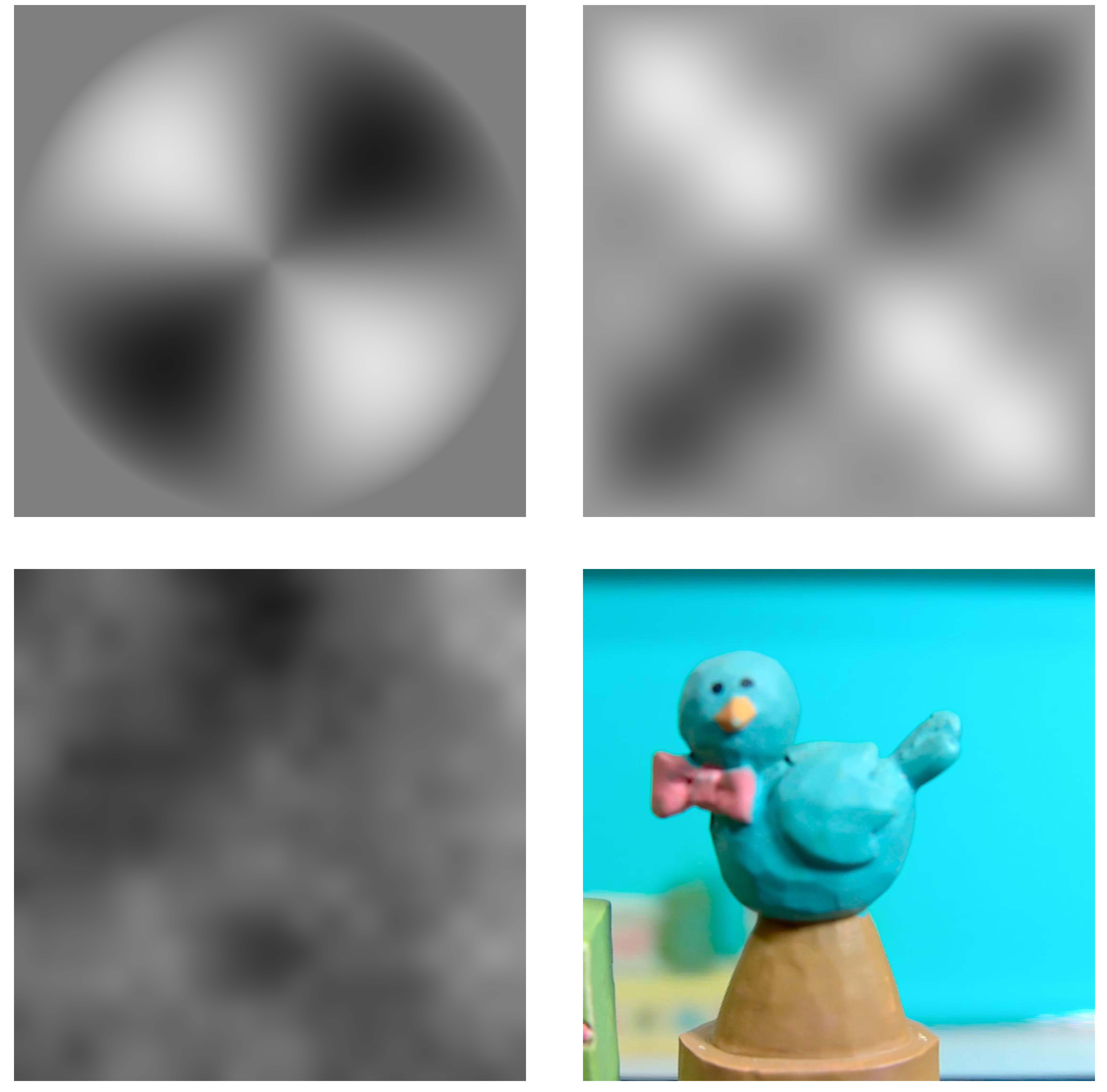}
	\caption{The four images that were used in  performance analysis. Top : patterns 1 and 2, bottom : patterns 3 and 4.}
	\label{fig:patterns}
\end{figure}

We tested our algorithm on a $3.7$ GHs AMD with a $16$ GB of memory. We implemented the results shown in Figures in Java and tested them on the Windows platform. We tested the running time of the algorithm against two parameters : changing number of slices in the covers and increasing the resolution of the image. The images that we used in our tests are shown in Figure \ref{fig:patterns}. See also Figure \ref{performance} for the performance analysis.

We also ran a comparison between the Mapper algorithm we present here and a contour tree algorithm. The contour tree algorithm we used is a version of algorithm given in \cite{carr2003computing11}. 

While both contour tree and Mapper give almost identical performance for images with small resolutions, Mapper outperforms contour tree as we increase the resolution of on the image. See Figure \ref{performance}.

One can notice here that the performance computation time of Mapper increase linearly with the increase of number of slices in the cover. Moreover, observe in  Figure \ref{performance} that Mapper computes faster than contour tree even when we choose to calculate it on the highest resolution. 
%\cite{doraiswamy2009efficient}.

\section{\uppercase{Limitations}}
\label{sec:limitation}

\begin{figure}[!h]
	\centering
	\includegraphics[width=0.95\linewidth]{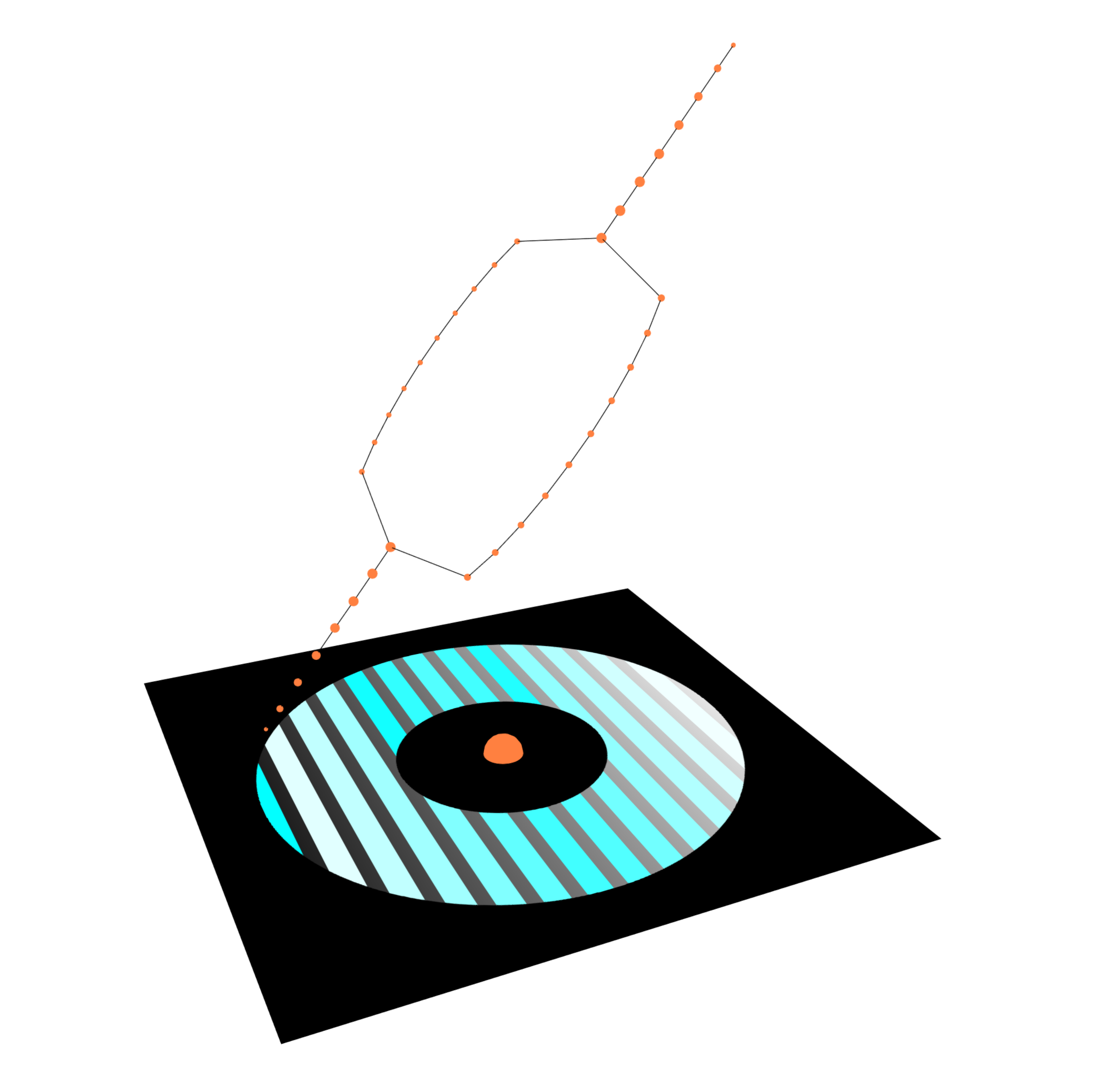}
	\caption{Mapper on discontinuous height function of an image is not guaranteed to produce a tree.}
	\label{fig:counter examples}
\end{figure}

Mapper assumes the underlying height function to be continuous. If the provided function is not continuous Mapper still produces a graph, but it is no longer guaranteed that this graph is a tree. Figure \ref{fig:counter examples} an example of an image whose height function is discontinuous. This image was created by drawing the ring shape with a constant single color then multiplying pixel values of this image by a gradient. The resulting graph has a clear cycle obtained by the nature of the discontinuity in the ring. 

\begin{figure}[!h]
	\centering
	\includegraphics[width=0.95\linewidth]{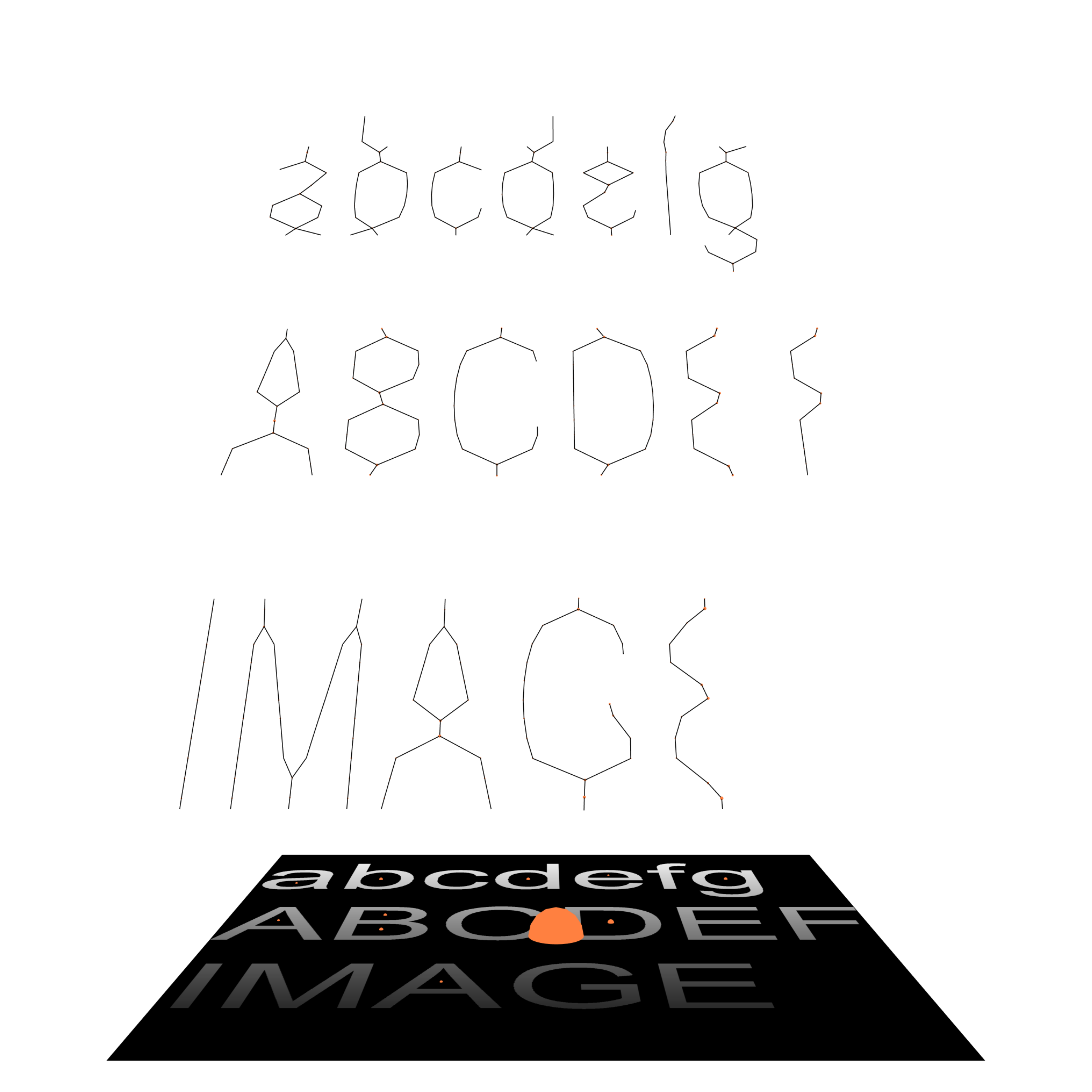}
	\caption{Mapper on an image consists of letters. The Mapper construction on this image gives a collection of disjoint graphs that have the same shape of as the letters written in the image.}
	\label{fig:counter examples2}
\end{figure}

Depending on the application at a hand this limitation of Mapper could potentially be used for image understanding. As illustrated in Figure \ref{fig:counter examples} the graph captures the "shape" in the underlying image. This is illustrated further in Figure \ref{fig:counter examples2}. 

\section{\uppercase{Conclusions and future work}}
\label{sec:conclusion}

Mapper is a powerful tool that can be used to study the topology of a certain domain with a scalar function attached to it. Mapper was originally defined and studied on point clouds.  We introduce the study of Mapper on simply connected domains and in particular 2d images. On simply connected domains, the Mapper construction generalizes contour, split, and join trees. The Mapper construction has multiple advantages over other contour tree algorithms. All previous contour tree algorithms assume the height function to be piecewise linear Morse functions. By assuming only continuity on the height function the Mapper graph allows us to extend the study of images using topology-based approaches to a much larger class of images. Most research related to Mapper has been geared towards topological Mapper. Our work here uses the properties of the image domain to obtain a customized algorithm for Mapper on images, which we show to have advantages in making the graph calculation more efficient. The algorithmic aspects to deal with additional domains have also been addressed in this work. We plan to investigate such directions more in the future.

\section*{\uppercase{Acknowledgements}}
\noindent This work was supported in part by a grants from the National Science Foundation (IIS-1513616) and (OAC-1443046).

\vfill
\bibliographystyle{apalike}
{\small
\bibliography{example}}
\vfill
\end{document}